\documentclass[10pt,twocolumn,letterpaper]{article}

\usepackage[pagenumbers]{cvpr} 

\usepackage{multirow}
\usepackage{makecell}
\usepackage{bbding}
\usepackage{bbm}
\usepackage{amssymb}
\usepackage{colortbl}
\usepackage{pifont}
\usepackage{xcolor}
\usepackage{tabularx}
\definecolor{SpringGreen4}{RGB}{46,139,87}
\definecolor{Honeydew4}{RGB}{131, 139, 131} 

\definecolor{cvprblue}{rgb}{0.21,0.49,0.74}

\usepackage[colorlinks, linkcolor=red,  anchorcolor=blue, citecolor=cvprblue]{hyperref} 

\title{HyperWalker: Dynamic Hypergraph-Based Deep Diagnosis for Multi-Hop Clinical Modeling across EHR and X-Ray in Medical VLMs}

\author{
Yuezhe Yang$^{1*}$, Hao Wang$^{1*}$, Yige Peng$^{1}$, Jinman Kim$^{2\dagger}$, Lei Bi$^{1\dagger}$ \\ 
$^{1}$Institute of Translational Medicine, Shanghai Jiao Tong University, Shanghai, China \\
$^{2}$School of Computer Science, University of Sydney, Sydney, Australia \\
$^{*}$Equal contribution \quad $^{\dagger}$Corresponding authors
}

\begin{document}
\maketitle
\begin{abstract}
    Automated clinical diagnosis remains a core challenge in medical AI, which usually requires models to integrate multi-modal data and reason across complex, case-specific contexts. Although recent methods have advanced medical report generation (MRG) and visual question answering (VQA) with medical vision-language models (VLMs), these methods, however, predominantly operate under a sample-isolated inference paradigm, as such processing cases independently without access to longitudinal electronic health records (EHRs) or structurally related patient examples. This paradigm limits reasoning to image-derived information alone, which ignores external complementary medical evidence for potentially more accurate diagnosis. To overcome this limitation, we propose \textbf{HyperWalker}, a \textit{Deep Diagnosis} framework that reformulates clinical reasoning via dynamic hypergraphs and test-time training. First, we construct a dynamic hypergraph, termed \textbf{iBrochure}, to model the structural heterogeneity of EHR data and implicit high-order associations among multimodal clinical information. Within this hypergraph, a reinforcement learning agent, \textbf{Walker}, navigates to and identifies optimal diagnostic paths. To ensure comprehensive coverage of diverse clinical characteristics in test samples, we incorporate a \textit{linger mechanism}, a multi-hop orthogonal retrieval strategy that iteratively selects clinically complementary neighborhood cases reflecting distinct clinical attributes. Experiments on MRG with MIMIC and medical VQA on EHRXQA demonstrate that HyperWalker achieves state-of-the-art performance. Code is available at: \url{https://github.com/Bean-Young/HyperWalker}
\end{abstract}

\section{Introduction}
Automated clinical diagnosis is a core challenge in intelligent healthcare. The underlying aim is to replicate the complex decision-making processes of clinicians by interpreting complex multimodal medical data via artificial intelligence (AI) systems  \cite{ma2025fully}. In this context, medical report generation (MRG) and medical visual question answering (VQA) have emerged as critical benchmarks. While MRG focuses on generating detailed diagnostic reports from medical images, medical VQA systems provide precise natural language answers to clinical queries. In real-world clinical practice, these tasks are inherently contextual: clinicians do not interpret imaging findings in isolation. Instead, they ground their diagnosis in longitudinal Electronic Health Records (EHRs) and make comparisons to structurally similar prior cases, and integrate relevant medical knowledge to ensure the diagnosis is accurate \cite{moor2023foundation}. Therefore, robust automated clinical diagnosis requires not only visual perception, but more rigorous, evidence-based reasoning capabilities. 

\begin{figure}
    \centering
    \includegraphics[width=\linewidth]{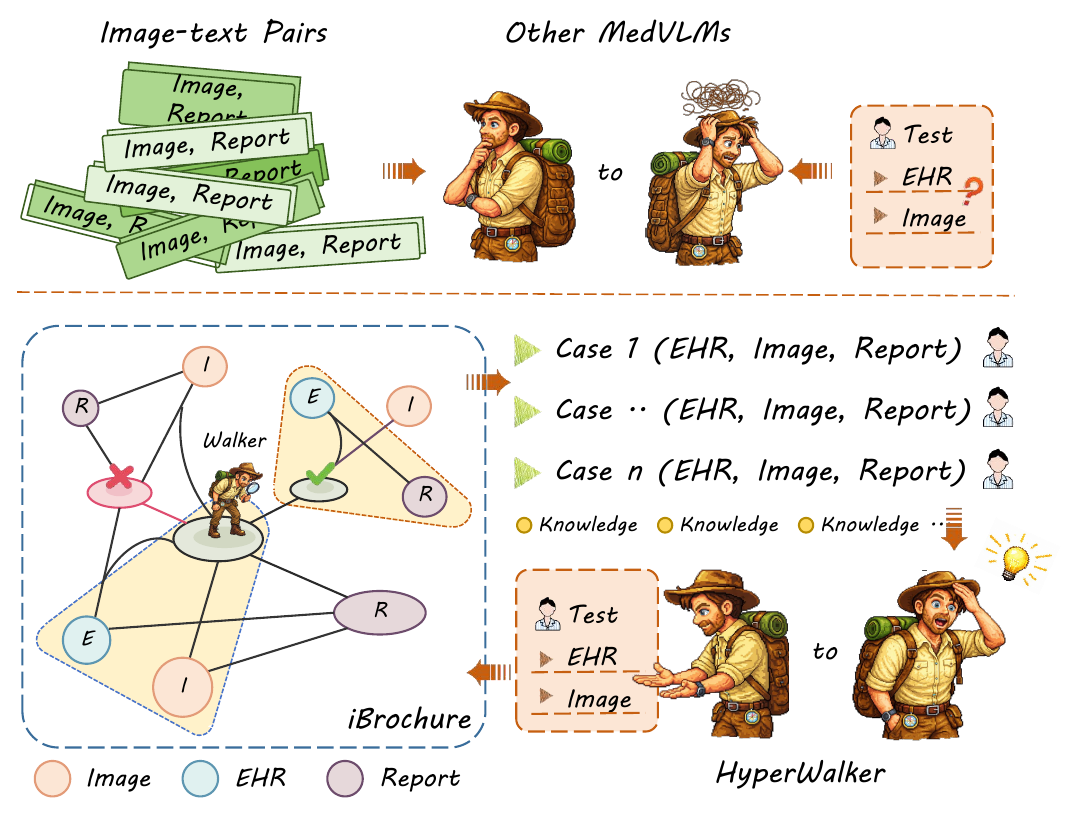}
    \caption{Conventional medical VLMs rely on sample-isolated inference, whereas HyperWalker performs structured multi-hop diagnostic reasoning over an implicit multimodal clinical hypergraph.}
    \label{fig:overview}
\end{figure}

Recent advances in medical vision-language models (VLMs) have shown promise in automated clinical diagnosis (e.g., MRG and VQA) by aligning visual and textual representations \cite{soni2025review,xu2025lingshu,yang2025auto}. These medical VLMs typically follow a two-stage fine-tuning paradigm: pre-training on large-scale medical image-text datasets to bridge the gap between visual and textual embeddings, followed by supervised fine-tuning (SFT) or reinforcement learning (RL) for specific tasks \cite{li2023llava,wu2025towards}. Although this paradigm has achieved state-of-the-art (SOTA) performance on MRG and medical VQA tasks, such simple alignment lacks the capacity for diagnostic reasoning and the effective application of medical knowledge. To address these limitations, some studies explicitly incorporate diagnostic reasoning into model outputs, while others attempt to inject medical knowledge through prompting strategies or retrieval-augmented systems \cite{pan2025medvlm,wang2025mrgagents}. For example, Li et al. \cite{li2025fine} propose to combine graph neural networks with VLMs, which transform Ocular images into bioinformatic graphs and then generate explanatory text to enable interpretable and precise staging of diabetic retinopathy. These strategies enhance explicit reasoning and knowledge utilization, thereby effectively improving overall diagnostic accuracy.

Despite these advances, existing methods heavily rely on isolated patient metadata while neglecting longitudinal EHRs during inference \cite{wu2025towards}. However, clinical diagnosis is essentially a continuous process where imaging findings must be interpreted within the context of the patient's medical history. When this context is absent, models are forced into a sample-isolated inference paradigm, treating each patient encounter as a static, independent event. This isolation restricts reasoning to decontextualized visual features and deprives the models of the comparative reasoning capabilities that human clinicians possess. In practice, experts rarely diagnose in isolation; rather, they ground their interpretations by correlating current findings to structurally similar prior cases. The absence of such an external reference mechanism makes current models prone to errors in real-world clinical environments where EHRs are lengthy, noisy, and structurally diverse.

Integrating longitudinal EHRs with VLMs for clinical reasoning presents three formidable challenges. First, the intrinsic structural heterogeneity of EHRs and their intricate interplay with medical images often elude traditional tree or binary graph representations, hindering the construction of a robust prior knowledge base. Second, the redundancy within EHRs prevents current algorithms from extracting diagnostic cues effectively, even when associations are established. Third, the prevalence of comorbidities and concurrent conditions requires models to move beyond single-case analysis and perform integrated reasoning across multi-sample scenarios to ensure diagnostic accuracy.

To address these fundamental limitations, we propose HyperWalker, shown in Fig.\ref{fig:overview}, a deep diagnosis framework that reformulates clinical reasoning into implicit multi-modal relationship modeling and diagnostic path modeling, combined with test-time adaptive learning.  First, we construct a dynamic heterogeneous hypergraph, termed iBrochure, to implicitly model the interconnected clusters of patient EHRs, images, and clinical attributes. To navigate the complex, high-dimensional memory space where simple similarity searches often fail, we introduce Walker, an RL agent. Guided by multi-dimensional rewards, Walker learns to traverse the hypergraph to identify evidence that balances clinical relevance, consistency, and diversity. Simultaneously, we introduce a linger mechanism that employs a multi-hop orthogonal retrieval strategy to aggregate complementary clinical facets, addressing the needs for comprehensive and diverse diagnosis. Finally, to mitigate shifts and satisfy the real-time constraints of clinical deployment, we implement a test-time training (TTT) mechanism. This lightweight adaptation enables HyperWalker to internalize case-specific nuances, effectively bridging the gap between generalized foundation knowledge and individualized clinical evidence. Our core innovations are summarized as follows:

\vspace{0.5em}

\noindent  \textbf{1)} We propose a deep diagnosis framework, termed HyperWalker, which, for the first time, breaks the sample-isolated inference paradigm by simulating real-world clinician logic where longitudinal EHRs, images, and medical knowledge are integrated via a recursive multi-hop reasoning process.

\noindent  \textbf{2)} We develop a heterogeneous medical hypergraph, iBrochure, to capture complex $n$-$ary$ clinical associations that are critical for cross-patient comparative reasoning. In addition, we design a RL–based optimizer, Walker, equipped with a linger mechanism to ensure diverse and information-rich complementary clinical diagnoses.

\noindent  \textbf{3)} Built upon this framework, we introduce a TTT mechanism that effectively calibrates the model’s output distribution for specific test cases, achieving SOTA performance on the MIMIC and EHRXQA datasets, and providing a feasible solution for building clinically deployable multimodal medical foundation models.

\section{Related Work}
\subsection{Medical VLMs}
Medical VLMs aim to extend the capabilities of general VLMs to accommodate medical-related tasks. These extended capabilities primarily include medical image-text alignment and improved medical diagnosis. Existing studies explore this adaptation through the pre-training stage and the post-training stage. In the pre-training stage, recent medical VLMs leverage large-scale image-text pretraining with contrastive, generative, or knowledge-guided objectives to learn joint representations. Contrastive learning approaches (e.g., ConVIRT \cite{zhang2022contrastive}, GLoRIA \cite{huang2021gloria}) align visual features with radiology report text by maximizing similarity for paired images and reports, yielding label-efficient representations that transfer well to retrieval and classification tasks. Furthermore, BioViL \cite{boecking2022making} and MedCLIP \cite{wang2022medclip} extend this paradigm by combining contrastive alignment with auxiliary generative losses to capture fine-grained clinical semantics. Additionally, knowledge-enhanced pretraining approaches (e.g., medKLIP \cite{wu2023medklip}) leverage domain-specific labels and ontologies to guide clinically meaningful image-text alignment.

After pretraining, the resulting medical VLMs are further adapted to specific tasks through SFT and reward-based alignment techniques. Standard SFT on curated datasets allows the model to specialize in producing accurate radiological answers and captions under direct supervision \cite{li2023llava}. Beyond supervised learning, RL can further refine outputs to maximize clinical correctness. CLARIFID \cite{lee2025clarifid} employs proximal policy optimization (PPO) \cite{schulman2017proximal} with a radiology-specific reward (CheXbert F1) to improve the factual accuracy of generated impressions. Furthermore, human preference alignment strategies \cite{wang2025mrgagents} have been explored to tune models with radiologist feedback, explicitly optimizing outputs to better match expert preferences and reduce hallucinations. However, current fine-tuning approaches of medical VLMs predominantly operate under a sample-isolated paradigm. They process each case as an independent instance, lacking the architectural mechanisms to integrate longitudinal EHRs or perform cross-patient comparative reasoning, which limits their diagnostic depth in complex clinical scenarios. 

\subsection{Hypergraph Theory in Multimodal Learning}
Hypergraphs have emerged as powerful tools for modeling complex semantic relationships in vision-language tasks, capturing higher-order associations beyond simple pairwise correspondences. Early approaches leveraged spectral hypergraph methods to learn joint image-text representations. Gao et al. \cite{gao2011tag} built a visual-text joint hypergraph for social image search, and Sparse Multimodal Hashing \cite{wu2013sparse} introduced a hypergraph Laplacian framework to preserve both intra-modal and inter-modal similarities in shared embeddings. With the development of deep learning, researchers extended hypergraph neural networks to multimodal data, enabling message passing among groups of features. Hypergraph Attention Networks \cite{chen2020hypergraph} align image and text features by constructing modality-specific hypergraphs and computing cross-hypergraph attention to match corresponding substructures. 

Hypergraph theory has also guided the development of advanced multimodal fusion and reasoning strategies. Hypergraph attention mechanisms allow models to attend to high-order groupings of visual and textual elements, improving fine-grained image-text retrieval. Yao et al. \cite{yao2022hypergraph} dynamically construct hyperedges among image regions and words to capture multi-way correspondences in remote sensing image-text retrieval. Similarly, Chen et al. \cite{chen2018supervised} introduce intra-modal and inter-modal hypergraphs to explicitly link image and text feature sets, combined with an adaptive fusion module to enrich image-text matching. In multimodal classification settings, heterogeneous hypergraphs enable richer fusion of modalities. HGMF \cite{chen2020hgmf} represents each data sample as a hypernode in a hypergraph and performs attention-based message passing over hyperedges to integrate incomplete image-text data. Other work embeds both modalities into unified hypergraph structures, thereby facilitating higher-order reasoning \cite{li2024variational}. Nevertheless, effective application of multimodal hypergraphs to medical VLM reasoning remains unexplored and requires a unified architecture that can model complex cross-modal relationships between longitudinal EHRs and sparse medical images.

\subsection{Multi-hop VLM in Deep Research}
To address complex queries that require stepwise inference across images and text, recent VLMs has moved toward multi-hop reasoning mechanisms. Several methods implement compositional reasoning by traversing intermediate entities or supporting facts between modalities. For example, general-domain frameworks like GraftNet \cite{liu2022graftnet} build reasoning chains that link visual content with external textual evidence or knowledge base entries, assembling question-specific subgraphs of image regions and facts. Similarly, the MM-GNN \cite{gao2020multi} model represents an image with multi-modal graphs and propagates messages between these sub-graphs in multiple steps, refining features through cross-modal context to answer complex questions. These approaches achieve improved visual question answering by performing iterative, multi-hop inference rather than relying on single-step attention, effectively combining disparate clues into coherent answers. In the medical domain, analogous VLMs \cite{liu2025aligning} are emerging that adopt similar multi-hop paradigms. They integrate medical knowledge graphs with imaging data to traverse explanatory links \cite{nguyen2025vehme} for robust diagnostic reasoning.

Despite the rapid progress of multi-hop VLMs, their application in the medical domain remains challenging, particularly when accounting for the structural heterogeneity of EHR data. By seamlessly integrating the strong representational capabilities of VLMs with the ability of hypergraphs to model complex multimodal relationships, and further incorporating TTT into a unified framework, HyperWalker presents the first multi-hop medical VLM designed for deep diagnosis.

\section{Methodology}

\begin{figure*}
    \centering
    \includegraphics[width=\textwidth]{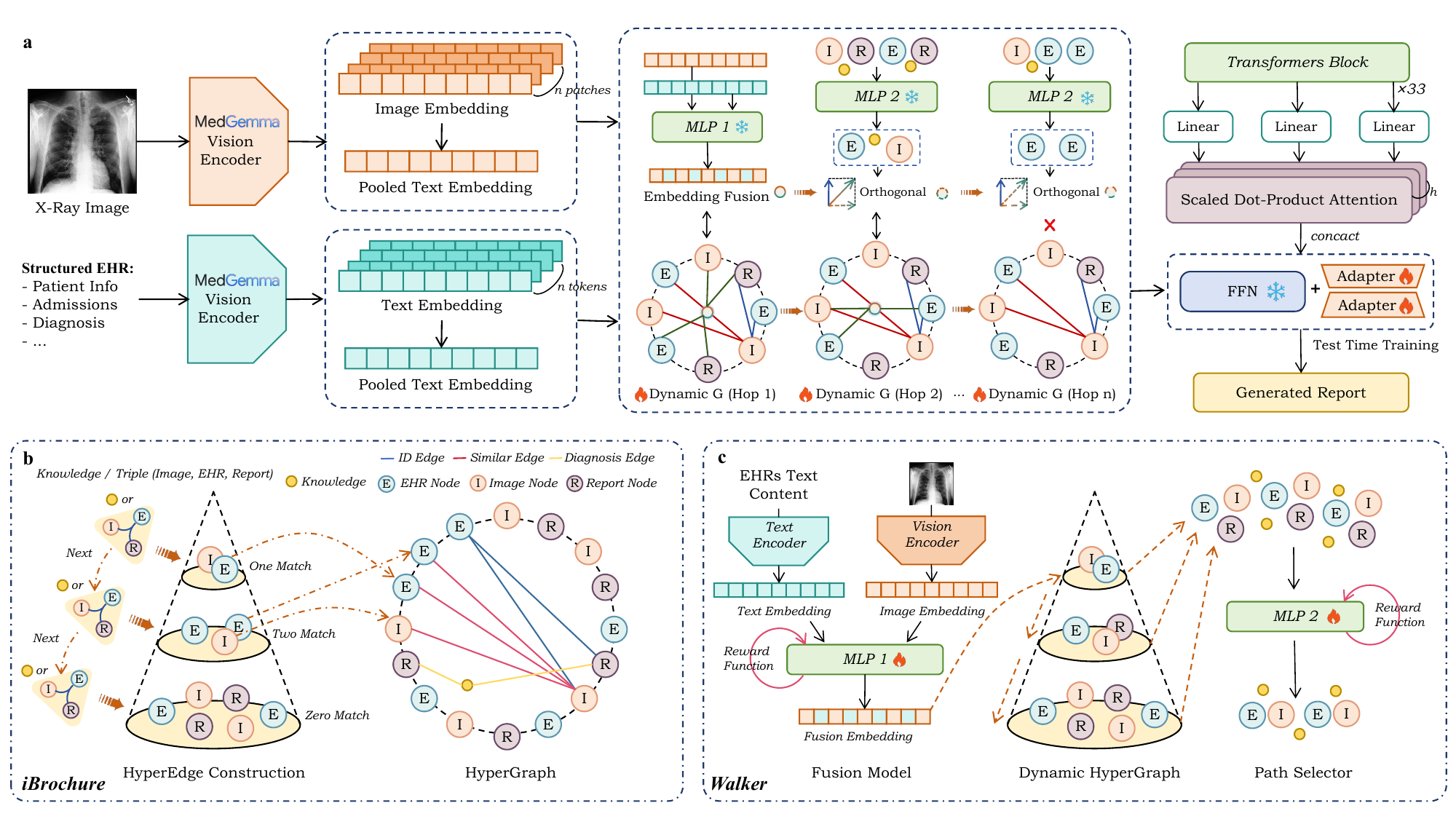}
    \caption{Architecture of HyperWalker, illustrating (a) multimodal encoding and fusion with dynamic orthogonal multi-hop reasoning, (b) implicit clinical hypergraph iBrochure construction via heterogeneous hyperedges, and (c) the reward-guided path selector Walker for modeling the diagnosis pathway.}
    \label{fig:flow}
    \vspace{-\baselineskip}
\end{figure*}

\subsection{Overview}

In this section, we provide an overview of HyperWalker as shown in Fig.\ref{fig:flow}. We reformulate the task of MRG and medical VQA as a recursive trajectory optimization problem over an implicit clinical manifold. The core of \textbf{HyperWalker} is divided into two synergistic components: \textbf{iBrochure}, which handles implicit multimodal relation modeling, and \textbf{Walker}, an autonomous agent for deep diagnosis path modeling.

We first define the medical multimodal hypergraph as $\mathcal{G} = (\mathcal{V}, \mathcal{E})$, where the vertex set $\mathcal{V}$ represents a heterogeneous collection of clinical entities including structured EHR information, X-ray image, radiology report and clinical diagnosis knowledge embeddings. A hyperedge $e \in \mathcal{E}$ is defined as a non-empty subset of $\mathcal{V}$ that encapsulates high-order, non-linear correlations between these disparate modalities. 

The objective of MRG and medical VQA is to find a textual sequence $Y$ that maximizes the conditional probability $P(Y \mid I, E, \mathcal{G})$, where $I$ and $E$ denote the input X-ray and EHR data, respectively. Similarly, medical question answering is defined as the generation of an optimal answer $A$ derived from the refined latent state of the clinical manifold.

The HyperWalker pipeline begins with iBrochure, which is responsible for constructing the implicit multimodal relationship manifold. Unlike traditional retrieval systems that rely on explicit data pairs, iBrochure projects EHR, images, reports, and expert knowledge into a shared latent embedding space provided by VLM encoders, and clinical relationships are modeled implicitly. To facilitate efficient discovery within this high-dimensional space, we implement a Hierarchical Navigable Small World (HNSW) index \cite{malkov2018efficient}, allowing for the rapid retrieval of relevant clinical subspaces in logarithmic time. This structure allows the knowledge repository to be a rapidly updatable, dynamic manifold that preserves complex medical dependencies.

Navigating this manifold is the Walker module, which implements autonomous path modeling via RL. Walker acts as an intelligent agent that simulates the iterative diagnostic logic of a physician by treating the hypergraph as a state space. At each step, the agent evaluates the current patient state and selects the most informative nodes to explore. This navigation is intrinsically coupled with a TTT mechanism, in which local adapters are fine-tuned for each specific case to refine the perception of the underlying vision–language model. To ensure the diversity of the reasoning trajectory, we introduce a linger mechanism, an orthogonal multi-hop strategy that encourages the agent to explore complementary clinical evidence spanning multiple candidate diagnoses. Through recursively refining multimodal embeddings via this autonomous navigation process, HyperWalker achieves deep clinical alignment before synthesizing the final diagnostic report or answer.

\subsection{iBrochure: Implicit Multimodal Relation Modeling}

The iBrochure module serves as the architectural foundation of HyperWalker, transforming fragmented clinical data into a coherent and searchable multimodal manifold. Rather than an explicit low-dimensional graph, iBrochure represents an implicit hypergraph $\mathcal{G} = (\mathcal{V}, \mathcal{E})$ where relationships are defined through high-dimensional geometric proximity and clinical logic.

\subsubsection{Multimodal Node Projection}

The construction process begins by projecting heterogeneous clinical entities, including EHR records, X-ray images, radiology reports and expert knowledge, into a unified latent space. For each clinical entity $v_i \in \mathcal{V}$, we utilize a modality-specific encoder $\Phi$ to generate a $D$-dimensional embedding $z_i = \Phi(v_i) \in \mathbb{R}^{1024}$. To mitigate the noise inherent in repeated clinical examinations and documentation, we implement a redundancy-aware pruning mechanism.
For any two EHR nodes $v_i, v_j$ within the same clinical study, their semantic similarity is computed via the cosine similarity function:
\begin{equation}
S(z_i, z_j) = \frac{z_i^\top z_j}{\|z_i\|_2 \|z_j\|_2} .
\end{equation}
If $S(z_i, z_j) > \tau_{prune}$, where $\tau_{prune} = 0.9$, the nodes are merged into a singular representative node through attentive aggregation, thereby ensuring the compactness of the manifold $\mathcal{M}$. This denoising step is critical for maintaining a salient clinical state space, preventing the subsequent navigation agent from being distracted by synonymous clinical descriptors.

\subsubsection{Hierarchical Manifold Indexing via HNSW}

To enable efficient diagnostic reasoning over large-scale longitudinal clinical data, iBrochure adopts a HNSW index to support fast retrieval on the implicit clinical manifold. The index is constructed over the metric space $(\mathcal{Z}, d)$, where the distance function $d(z_i, z_j) = 1 - S(z_i, z_j)$ measures semantic divergence between multimodal clinical entities in the shared latent space.

HNSW organizes the manifold into multiple hierarchical layers with sparse and adaptive connectivity, enabling logarithmic-time approximate nearest neighbor search. This hierarchical structure allows the model to rapidly localize clinically relevant regions of the manifold without exhaustive traversal of the global hypergraph, which is critical for real-time diagnostic reasoning.

During inference, given a query state $z_q$ derived from the current patient context, the HNSW index retrieves a clinically relevant subspace $\mathcal{V}_{sub} \subset \mathcal{V}$. This retrieved subspace provides a focused diagnostic context for subsequent multi-hop reasoning by the Walker agent. By decoupling global manifold growth from local diagnostic retrieval, the proposed indexing strategy ensures scalable and responsive clinical reasoning as the hypergraph continuously evolves with new longitudinal patient data.

\subsubsection{Implicit Hyperedge Induction}

The core innovation of iBrochure lies in its implicit hyperedge induction mechanism, which captures high-order correlations across disparate modalities. We define the incidence matrix $H \in \mathbb{R}^{|\mathcal{V}| \times |\mathcal{E}|}$, where each entry $h_{i,j}$ denotes the strength of the association between node $v_i$ and hyperedge $e_j$. Unlike explicit graphs, our hyperedges are induced through four distinct logic layers based on global manifold thresholds:

First, id-based hyperedges establish temporal consistency by connecting all nodes $v_i$ that share a common study identifier. Second, similarity-based hyperedges are dynamically formed between EHR and image nodes. Diverging from local neighborhood constraints, we define a similarity-based hyperedge $e_{sim}$ for each node $v_i$ by capturing all multimodal neighbors within the clinical manifold that satisfy a global similarity threshold $\tau_{sim} = 0.8$:
\begin{equation}
e_{sim}(v_i) = \{ v_j \mid S(z_i, z_j) > \tau_{sim}, v_j \in \mathcal{V}_{ehr} \cup \mathcal{V}_{img} \} .
\end{equation}

Third, to capture cross-case semantic consistencies, we define diagnosis hyperedges $e_{rep}$ that connect report nodes $v_r$ to other historical reports $v_m$ within the manifold. Similarly, this edge is constructed by capturing all relevant semantic neighbors that satisfy the clinical similarity threshold:
\begin{equation}
e_{rep}(v_r) = \{ v_m \mid S(z_r, z_m) > \tau_{sim}, v_m \in \mathcal{V}_{report} \} .
\end{equation}

Finally, disease-based hyperedges bridge clinical findings with formal medical logic. Every report node $v_r$ is associated with a knowledge anchor $v_k$ by maximizing the disease-based alignment:
\begin{equation}
e_{dis}(v_r) = \text{argmax}_{v_k \in \mathcal{V}_{kb}} S(z_r, z_k) .
\end{equation}
By programmatically tethering each report to the knowledge anchor with the highest similarity score, we ensure that each report is connected to its most important clinical knowledge node. This hierarchical topology allows iBrochure to model an implicit clinical manifold that supports deep, evidence-based diagnostic navigation.

\subsection{Walker: Deep Diagnosis through Multi-hop Path Modeling}

The Walker module constitutes the decision-making core of the HyperWalker framework, functioning as an autonomous agent that navigates the iBrochure manifold to synthesize complex clinical evidence. By integrating multimodal fusion, RL selection, and recursive test-time training, Walker effectively simulates the iterative diagnostic workflow of human experts.

\subsubsection{Multimodal Fusion}

To initiate the reasoning process, Walker first maps the disparate input modalities into a unified query state through a Feature-wise Linear Modulation strategy. Given the initial image embedding $z_{img}$ and the electronic health record embedding $z_{ehr}$, the fusion module employs a multi-layer perceptron to derive scaling parameters $\gamma$ and shift parameters $\beta$ from the EHR context. The modulated multimodal representation is formulated as:
\begin{equation}
z_{fused} = \text{LN}(z_{img} + (\gamma \odot z_{img} + \beta)) ,
\end{equation}
where LN denotes Layer Normalization. To ensure that the fused query remains within the metric space of the iBrochure manifold, we apply L2 normalization to project $z_{fused}$ onto a unit hypersphere, maintaining compatibility with the HNSW cosine index. A crucial design element is the identity initialization of the modulation layers, which ensures that the initial query preserves the fundamental visual features while gradually incorporating EHR-driven context during training.

\subsubsection{Autonomous Path Selection based on RL}

Path modeling within the clinical manifold is governed by an RL-based selector that evaluates candidate nodes retrieved from the hypergraph. The selector employs a policy network to score each candidate node $v_i$ by jointly encoding the current query context and node representation. Specifically, given a query embedding $z_q$ and candidate node embeddings $\{z_i\}$, the policy induces a stochastic selection distribution
\begin{equation}
P(v_i \mid z_q) = \frac{\exp\!\left(\pi_\theta([z_q, z_i]) / T\right)}{\sum_j \exp\!\left(\pi_\theta([z_q, z_j]) / T\right)},
\end{equation}
where $\pi_\theta(\cdot)$ denotes a learnable scoring function and $T=0.01$ is a temperature parameter that sharpens the policy, enabling the agent to more decisively focus on diagnostically salient clinical anchors.

The policy optimization is driven by a set of complementary reward components designed to balance semantic relevance, diagnostic coverage, and reasoning efficiency. The accuracy reward $R_{\mathrm{acc}}$ encourages semantic alignment between the query and the selected evidence set $\mathcal{S}$ by measuring their cosine similarity:
\begin{equation}
\mathcal{R}_{\mathrm{acc}} = \frac{1}{|\mathcal{S}|} \sum_{v_i \in \mathcal{S}} \cos(z_q, z_i).
\end{equation}

To promote comprehensive diagnostic reasoning across multiple disease hypotheses, the diversity reward $R_{\mathrm{div}}$ penalizes redundancy among selected nodes by discouraging overly similar evidence:
\begin{equation}
\mathcal{R}_{\mathrm{div}} = 1 - \frac{1}{|\mathcal{S}|(|\mathcal{S}|-1)} \sum_{i \neq j} \cos(z_i, z_j).
\end{equation}

Reasoning efficiency is enforced through penalties on both the traversal depth and the number of hops in the reasoning trajectory, defined as
\begin{equation}
\mathcal{R}_{\mathrm{dp}} = \frac{d}{d_{\max}} ,  \mathcal{R}_{\mathrm{hp}} =\frac{h}{h_{\max}},
\end{equation}
where $d$ denotes the current reasoning depth and $h$ represents the hop count. $d_{\max}=5$ and $h_{\max}=5$ are pre-defined maximum reasoning budgets in the current setting, which are used to normalize the penalties on reasoning depth and hop count.

Finally, the overall reward function is formulated as a weighted combination of the above components:
\begin{equation}
\mathcal{R} = \lambda_a \mathcal{R}_{\mathrm{acc}} + \lambda_d \mathcal{R}_{\mathrm{div}} - \lambda_p \left( \mathcal{R}_{\mathrm{dp}} + \mathcal{R}_{\mathrm{hp}} \right),
\end{equation}
where $\lambda_a=1$, $\lambda_d=0.5$, and $\lambda_p=0.3$ control the trade-off between accuracy, diversity, and efficiency.



\subsubsection{Linger Mechanism with TTT}

During the reasoning process, HyperWalker performs recursive TTT to dynamically calibrate the underlying VLM to patient-specific clinical characteristics. For each diagnostic case, the model fine-tunes lightweight local adapters embedded in the FFN using clinical triplets retrieved from the hypergraph constructed from the training set. This optimization is driven solely by the report generation objective, allowing the model to transition from a globally trained representation to a case-conditioned diagnostic state that captures localized clinical patterns within the retrieved clinical subspace.

To further enhance the exploratory capacity of the Walker agent and mitigate myopic evidence accumulation, we introduce the \emph{linger} mechanism, an orthogonal multi-hop reasoning strategy that explicitly promotes complementary diagnostic exploration. Rather than repeatedly traversing semantically redundant regions of the clinical manifold, the linger mechanism encourages the agent to probe alternative diagnostic directions. Given the current query representation $z_q$ and the mean embedding $\bar{z}_{\mathrm{sel}}$ of previously selected nodes, an orthogonalized query vector is computed as
\begin{equation}
z_{\mathrm{orth}} = z_q - \frac{z_q^\top \bar{z}_{\mathrm{sel}}}{\|\bar{z}_{\mathrm{sel}}\|^2} \bar{z}_{\mathrm{sel}}.
\end{equation}
This orthogonal component is subsequently used to guide further retrieval and reasoning steps, effectively steering the agent toward unexplored yet diagnostically relevant regions of the clinical manifold. By coupling recursive TTT with orthogonalized exploration, HyperWalker constructs a diverse and evidence-complete reasoning trajectory, enabling robust synthesis of the final diagnostic report or answer. The implementation details of TTT can be found in the \textbf{supplementary materials}.

\begin{table*}[!t]
  \centering

  \resizebox{0.95\textwidth}{!}{%
  {
    \setlength{\doublerulesep}{1.5pt}
    \setlength{\tabcolsep}{4pt}
    \setlength{\extrarowheight}{1.5pt}
    \renewcommand{\arraystretch}{1.0}

    \newcolumntype{S}{>{\centering\arraybackslash}p{1.5cm}}
    \newcolumntype{M}{>{\raggedright\arraybackslash}X}

    \begin{tabularx}{\textwidth}{M|S|SSSS|SSS}
      \hline\hline
      \textbf{Models} & w/ EHR & BL-1 & BL-4 & MTR & RG-L & Pre & Recall & F1 \\
      \hline

    \multirow{2}{*}{Llama3.2-Vision-11B} &
    \ding{55} & 10.21 & 0.70 & 7.09 & 11.30 & 16.12 & 14.87 & 15.03 \\
    &
    \ding{51} & 7.45 & 0.70 & 11.17 & 8.55 & 14.78 & 17.92 & 13.41 \\
    \cline{2-9}

      \multirow{2}{*}{Qwen3-VL} &
      \ding{55} & 14.72 & 1.74 & 18.81 & 14.11 & 12.99 & 17.60 & 13.39\\
      &
      \ding{51} & 11.55 & 1.35 & 18.37 & 11.78 & 30.27 & 20.36 & 20.99  \\
      \cline{2-9}

      \multirow{2}{*}{Gemma} &
      \ding{55} & 12.93 & 1.48 & 18.18 & 12.54 & 15.03 & 14.69 & 12.50 \\
      &
      \ding{51} & 13.17 & 1.33 & 17.07 & 12.35 & \textbf{32.51} & 20.97 & 20.51\\
      \hline

      \multirow{2}{*}{Lingshu} &
      \ding{55} & 19.99 & 3.82 & 16.70 & 19.81 & 21.40 & 24.16 & 21.56 \\
      &
      \ding{51} & 18.18 & 3.32 & 15.92 & 18.63 & 19.30 & 27.91 & 20.02 \\
      \cline{2-9}

      \multirow{2}{*}{MedGemma} &
        \ding{55} & 20.41 & 2.08 & 17.23 & 17.99 & 19.95 & 22.10 & 20.68 \\
      &
        \ding{51} &19.83 &1.67&17.53 & 16.62& 22.18 & 25.40 & 19.69  \\

      \hline
    \textbf{Ours} & \ding{51} & \textbf{25.78} & \textbf{4.98} & \textbf{19.32} & \textbf{22.47} & 26.15 & \textbf{30.88} & \textbf{29.69 }\\ \hline\hline

    \end{tabularx}
  }
  }
  \caption{Quantitative comparison of report generation performance under different modalities.}

  \label{tab:report_metrics}
   \vspace{-\baselineskip}
\end{table*}

\begin{table}[t]
  \centering
  \footnotesize
  \renewcommand{\arraystretch}{1.3}
  \begin{tabularx}{\linewidth}{
    >{\centering\arraybackslash}X|cc|cc}
    \hline\hline
    \multirow{2}{*}{\textbf{Models}} &
    \multicolumn{2}{c|}{\textbf{Generation}} &
    \multicolumn{2}{c}{\textbf{Model}} \\
    \cline{2-5}
    & BL-4 & F1 & Param.(B) & Inf.(s)\\
    \hline
    Qwen3-VL-Thinking & 0.70 & 23.72 & 7 & 86.56\\
    GLM-4.1V-Thinking & 0.76 & 24.26 & 9 & 51.82\\
    \hline
    MedVLM-R1         & 0.70 & 11.59 & $\approx$7& 12.31 \\
    LVMed-$\rm R^2$     & 1.43 & 20.08 & $\approx$7 & 60.62 \\
    \hline
    Ours              & \textbf{4.98}   & \textbf{29.69}   & \textbf{$\approx$4}  &  \textbf{4.66}   \\
    \hline\hline
  \end{tabularx}
  \caption{Comparison of reasoning performance and generation metrics across different VLMs.}
  \label{tab:ablation_limited}
\end{table}

\begin{table}[t]
  \centering
  \footnotesize
  {
    \renewcommand{\arraystretch}{1.3}
    \begin{tabularx}{\linewidth}{
      >{\centering\arraybackslash}X|c|ccc}
      \hline\hline
      \multirow{2}{*}{\textbf{Models}} &
      \multicolumn{1}{c|}{\textbf{True/False}} &
      \multicolumn{3}{c}{\textbf{Open-ended}} \\
      \cline{2-5}
      & Acc & $S_{Q}$ & $S_{G}$ & MTR \\
      \hline
      Llama3.2-Vision & 48.75 & 1.68 & 2.61 & 2.15 \\
      Qwen3-VL        & 58.33 & 1.75 & 2.93 & 1.78 \\
      Gemma           & 49.19 & 2.09 & 3.10 & 2.76 \\
      \hline
      Lingshu         & 60.77 & 2.18 & 3.20 & 4.17 \\
      MedGemma        & 52.54 & 1.94 & 2.77 & 3.40 \\
      \hline
      Ours & 
      \textbf{70.43} & \textbf{2.69} &\textbf{3.55} &\textbf{6.86} \\
      \hline\hline
    \end{tabularx}
  }
  \caption{VQA performance comparison across different models.}
  \label{tab:vqa_metrics}
  \vspace{-10pt}
\end{table}

\begin{table}[t]
  \centering
  \footnotesize
  {
    \renewcommand{\arraystretch}{1.3}
    \begin{tabularx}{\linewidth}{
      >{\centering\arraybackslash}c|
      >{\centering\arraybackslash}X|
      c|c|c}
      \hline\hline
      \textbf{Type} & \textbf{Variant} & BL-4 & F1 & Inf.(s) \\
      \hline
      \multirow{3}{*}{\textbf{iBrochure}} 
        & w/o EHR        & 3.62 & 22.41 & \textbf{4.64} \\
        & w/o X-Ray     & 1.15 & 12.86 & 4.87 \\
        & w/o Knowledge & 4.41 & 27.73 & 4.66 \\
      \hline
      \multirow{3}{*}{\textbf{Walker}} 
        & w/o $\mathcal{R}_{\mathrm{acc}}$ 
             & 3.78 & 23.56 & 5.07 \\
        & w/o $\mathcal{R}_{\mathrm{div}}$ 
             & 4.12 & 25.84 & 4.67 \\
        & w/o $\mathcal{R}_{\mathrm{dp}} \,\&\, \mathcal{R}_{\mathrm{hp}}$ 
             & 4.37 & 26.91 & 8.48 \\
      \hline
      \multicolumn{2}{c|}{\textbf{Full}} 
        & \textbf{4.98} & \textbf{29.69} & 4.66 \\
      \hline\hline
    \end{tabularx}
  }
  \caption{Ablation study on MRG.}
  \label{tab:ablation}
\end{table}

\section{Experiments}

\subsection{Datasets}

We evaluate HyperWalker on two multimodal benchmarks targeting distinct clinical tasks. For MRG, we use the MIMIC benchmark, which combines MIMIC-CXR \cite{johnson2019mimic} and MIMIC-IV \cite{johnson2023mimic} to provide paired chest X-ray images and longitudinal EHRs. The Findings and Impression sections of radiology reports are used as generation targets. For medical VQA, we adopt EHRXQA \cite{bae2023ehrxqa}, a multimodal benchmark that requires joint reasoning over structured EHR data and chest X-ray images. Dataset preprocessing and task-specific construction details are provided in the \textbf{supplementary materials}. 

\subsection{Experimental Setup}
We compare HyperWalker with a broad set of SOTA VLMs, including general-purpose models such as Llama3.2-Vision-11B \cite{meta2024llama}, Qwen3-VL \cite{yang2025qwen3}, and Gemma \cite{team2025gemma}, as well as medical-domain models like Lingshu \cite{xu2025lingshu} and MedGemma \cite{sellergren2025medgemma}. To assess complex reasoning, we include thinking-oriented models Qwen3-VL-Thinking \cite{Qwen3-VL} and GLM-4.1V-Thinking \cite{hong2025glm}, along with recent medical reasoning frameworks MedVLM-R1 \cite{pan2025medvlm} and LVMed-$\mathrm{R}^2$ \cite{wang2025lvmed}.

Report generation is evaluated using BLEU-1 (BL-1), BLEU-4 (BL-4), METEOR (MTR), ROUGE-L (RG-L), and clinical factuality metrics Precision, Recall, and F1. We also report model size (Para.) and inference time (Inf.) to compare computational efficiency. For medical VQA, binary questions are evaluated with accuracy, while open-ended questions use subjective scores $S_Q$ and $S_G$ assigned by Qwen- and Gemma-based evaluators, respectively, ranging from 1 to 5, with higher values indicating better quality, followed \cite{li2024llms}. MTR is additionally reported for lexical overlap. All experiments are implemented in PyTorch 2.8 with Transformers and run on an NVIDIA RTX PRO 6000 GPU with 96GB memory.

\subsection{Quantitative Evaluation}

The quantitative results for MRG in MIMIC, summarized in Table \ref{tab:report_metrics}, demonstrate that HyperWalker significantly outperforms both general and specialized baselines. HyperWalker achieves a BLEU-4 of 4.98 and an F1-score of 29.69, which represents a substantial improvement over MedGemma’s F1-score of 19.69. An important observation is the model's resilience to data heterogeneity: while most baselines exhibit performance degradation when naively augmented with longitudinal EHR data, HyperWalker effectively leverages the Walker agent to distill relevant multimodal evidence, confirming the necessity of active graph-based filtering over passive concatenation.

In terms of reasoning performance and inference efficiency, as detailed in Table \ref{tab:ablation_limited}, HyperWalker exhibits advantages. Despite its relatively compact parameter size ($\approx$ 4B), our model surpasses much larger thinking models in generation quality. While models like Qwen3-VL-Thinking require 86.56s for inference, HyperWalker completes the diagnostic trajectory in only 4.66s. This efficiency-accuracy trade-off proves that the recursive hypergraph navigation effectively replaces the need for massive chain-of-thought token generation. By "walking" to the answer rather than "hallucinating" a reasoning path, HyperWalker avoids the computational overhead of long-context generation. Furthermore, the VQA results on EHRXQA in Table \ref{tab:vqa_metrics} show that HyperWalker reaches an accuracy of 70.43\%, outstripping the next best medical model, Lingshu, by nearly 10\%. The high METEOR score of 6.86 in open-ended VQA highlights the model's ability to synthesize complex clinical evidence into precise natural language answers. Discussion and case study can be found in the \textbf{supplementary materials}.

\subsection{Ablation Study}

Table \ref{tab:ablation} presents ablation results for medical report generation. Removing individual modalities from iBrochure consistently degrades performance, with the absence of X-ray inputs causing the most severe drop, indicating that visual evidence is indispensable for accurate report generation. Excluding knowledge nodes also leads to noticeable declines in both BL-4 and F1, highlighting their role in maintaining clinical consistency.

For the Walker module, removing any reward component results in performance degradation, demonstrating that each objective contributes to effective reasoning. In particular, disabling diversity-related rewards leads to reduced factual accuracy, while removing depth and hop-related rewards substantially increases inference time, suggesting less efficient exploration. The full model achieves the best overall performance with competitive inference cost, confirming the complementary roles of iBrochure and Walker in HyperWalker.

\section{Conclusion}

This work addresses the limitations of the sample-isolated inference paradigm in existing medical VLMs, where clinical diagnosis is reduced to independent multimodal predictions detached from longitudinal context and comparative evidence. HyperWalker reframes diagnosis as an evidence-driven reasoning process by organizing heterogeneous clinical data into an implicit hypergraph, enabling structured interaction among imaging findings, EHR information, and historically related cases. Through controlled multi-hop path modeling, the framework progressively accumulates and refines diagnostic evidence, while the linger strategy promotes exploration of complementary clinical attributes. In addition, TTT further aligns model behavior with the local clinical context encountered during reasoning. Together, these mechanisms enable HyperWalker to capture high-order and cross-patient associations beyond sample-level modeling, leading to more accurate and clinically grounded generation across medical report writing and multimodal question answering.

\bibliographystyle{plain}
\bibliography{references}
\newpage
\appendix
\section*{Supplementary Material}  
\section{Test-Time Training (TTT) Procedure}

The core of HyperWalker's reasoning capability lies in the Deep Diagnosis mechanism, a recursive TTT protocol that calibrates the model's latent perception to the specific clinical context of a patient prior to inference. This section details the optimization protocol, the safeguards against overfitting, and the measures taken to ensure data integrity and clinical fairness. For medical report generation (MRG), the TTT workflow begins with the Multi-Relational Graph construction: 1\% of the training data is randomly sampled to construct the hypergraph, while another 1\% of data is used to train the reinforcement learning agent, Walker. During testing, for each individual test sample, the pre-built hypergraph iBrochure and the trained Walker agent are used to model the clinical diagnostic path. Specifically, raw embeddings are attached to the hypergraph without pooling, and matched triplets are recalled to optimize the adapters. Finally, the optimized adapters, together with the original vision-language model (VLM), generate the final diagnostic report. That is, HyperWalker optimizes solely with TTT for a single test sample. For the Medical Visual Question Answering (VQA) task, since the EHRXQA dataset is also derived from the MIMIC dataset, the workflow remains unchanged, with only the final response prompt being adapted to the question-answering format.

\subsection{Optimization Protocol and Hyperparameters}

The TTT process is triggered once the Walker agent identifies a set of salient diagnostic anchors within the iBrochure manifold. For each retrieved clinical triplet—comprising a historical image, its associated radiology report, and structured EHR data—a targeted weight update is performed on the model's internal adapters. To maintain computational efficiency suitable for real-time clinical deployment, a single-step gradient update is applied per diagnostic instance. We employ the AdamW optimizer with a highly conservative learning rate of $\eta = 10^{-5}$. The optimization objective is defined by the cross-entropy loss between the model's generated output and the ground-truth reports of the retrieved historical triplets, rather than the target report of the current test case. This ensures that the model "pre-activates" relevant medical concepts and vocabulary specific to the observed pathology before committing to a final diagnosis. Formally, the optimization can be expressed as:

\begin{equation}
\mathcal{L}_{TTT} = - \sum_{(x_i, y_i) \in \mathcal{T}} y_i \log \hat{y}_i, 
\quad \hat{y}_i = f_\theta(x_i; \phi)
\end{equation}

where $\mathcal{T}$ denotes the set of retrieved historical triplets, $f_\theta$ is the frozen backbone VLM, and $\phi$ represents the parameters of the adapters being optimized.

\subsection{Safeguards Against Overfitting}

A primary concern in test-time training is the risk of the model overfitting to a singular, potentially noisy test instance, which could lead to "catastrophic forgetting" of global medical knowledge. HyperWalker mitigates this through strict architectural constraints. During the Deep Diagnosis phase, the entire backbone of the VLM remains frozen, and only the low-rank adaptation (LoRA) layers or specialized adapters are updated. This preserves the model's fundamental linguistic and visual representations. Furthermore, the optimization is limited to a single gradient step, and a residual connection within the fusion module ensures that the model adjusts its "clinical focus" without deviating from the established diagnostic manifold.

\subsection{Data Integrity and Leakage Prevention}

Ensuring data integrity and preventing leakage are critical for fair evaluation. Our TTT protocol strictly enforces temporal and categorical isolation: only historical samples with different IDs from the training set are retrieved in the iBrochure manifold, while the data of the current test sample is entirely withheld during TTT. Moreover, we store embeddings in an implicit form rather than explicit textual nodes, making them non-invertible and resistant to unintended information leakage. This guarantees that the model's adaptation relies solely on clinically relevant historical patterns without exposure to the test instance itself, thereby maintaining both methodological rigor and patient-level fairness.

\section{Data Processing}
\subsection{MIMIC Dataset for MRG}

To construct the high-order clinical manifold required by the HyperWalker framework, we perform a comprehensive multimodal alignment between the MIMIC-IV clinical database and the MIMIC-CXR imaging database. This process ensures that each radiographic examination is accompanied by rich and temporally relevant clinical context. The data processing pipeline consists of three main stages: multimodal intersection matching, semantic enrichment and denoising, and precise temporal windowing.

During the multimodal intersection matching stage, we associate Electronic Health Records (EHRs) with their corresponding images and reports. Strict inclusion criteria are applied, retaining only subjects who possess both valid EHR entries and radiographic images. As a result, the pipeline curates a dataset comprising 56,845 unique subjects, while 3,243 subjects are excluded due to missing modality components.

The semantic enrichment and denoising stage involves relational joins between raw clinical event tables and their corresponding metadata dictionaries. Specifically, we leverage the \texttt{d\_icd\_diagnoses} and \texttt{d\_icd\_procedures} modules to map ICD-9 and ICD-10 codes to their canonical clinical descriptions. Similarly, laboratory results and intensive care unit observations are enriched by linking \texttt{itemid} fields to human-readable labels using the \texttt{d\_labitems} and \texttt{d\_items} catalogs. This transformation converts sparse numeric codes into semantically dense clinical descriptors. To further minimize input noise, we apply a deduplication strategy that removes redundant diagnostic and prescription records and strips purely numeric identifiers, thereby prioritizing meaningful semantic text.

To ensure the temporal relevance of clinical evidence, we adopt a 90-day temporal windowing strategy. Using the \texttt{StudyDate} and \texttt{StudyTime} fields from the MIMIC-CXR metadata, we compute an exact timestamp for each imaging study. We then filter the patient EHR history to retain only clinical events occurring within the 90 days preceding image acquisition. This temporal alignment is designed to reflect the diagnostic practice of radiologists, who typically focus on a patient’s recent medical history.

The final processed data are organized by subject identifiers and examination identifiers, and serialized into structured JSON files. A custom encoder is employed to remove null values and stabilize floating-point precision to \texttt{float16}, providing high-quality, low-noise inputs for constructing the iBrochure hypergraph.

\begin{figure}
    \centering
    \includegraphics[width=0.8\linewidth]{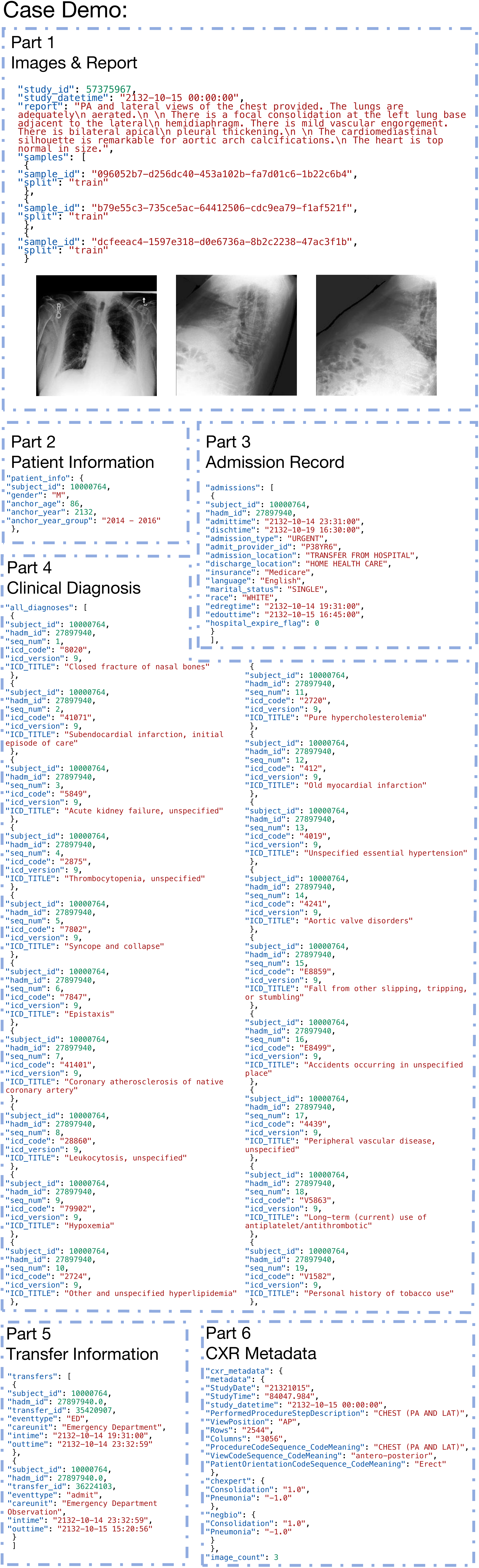}
    \caption{Case Demo for MIMIC}
    \label{fig:demo}
\end{figure}

Regarding data splits, since the proportion of filtered-out samples is relatively small, we preserve the original training, validation, and test splits provided by the MIMIC-CXR dataset when partitioning the processed data. Fig.\ref{fig:demo} illustrates a representative sample from the MIMIC training set, which includes longitudinal EHR records, the corresponding chest X-ray images, and the associated radiology report.
\subsection{EHRXQA for Medical VQA}

The EHRXQA dataset is used to evaluate the multimodal medical VQA capabilities of HyperWalker. For a fair comparison and to align with the training attributes available at test time, we exclusively use the gold test split, which contains expert-verified question–answer pairs. The initial test set consists of 4,808 samples, which are subjected to a rigorous multi-stage filtering process to ensure that the evaluation focuses on independent and genuinely multimodal clinical reasoning.

To enable a fair and deterministic assessment of the model’s diagnostic reasoning ability, the original gold test set is refined into a subset of 2,363 samples. The exclusion criteria are defined as follows. First, we remove 258 samples with null answers. Although null answers are included in the original dataset to increase answer entropy and reduce random guessing, we exclude them to focus on the model’s ability to correctly identify and describe the presence of specific clinical findings. Second, the EHRXQA dataset is constructed by integrating two unimodal resources, resulting in questions that may require either unimodal or cross-modal reasoning. To specifically evaluate cross-modal reasoning, we remove 1,716 samples that lack the \texttt{func\_vqa} operator. These queries can be resolved using unimodal textual information alone and do not require the joint reasoning over visual and EHR modalities that our framework is designed to test. Third, 334 samples are discarded because they lack explicit \texttt{patient\_id} or \texttt{study\_id} markers. Such samples often correspond to non-independent queries that rely on prior conversational context, making them unsuitable for evaluating autonomous diagnostic path modeling. Finally, we cross-reference all remaining samples with our processed MIMIC multimodal dataset, excluding 137 samples for which the required data are unavailable. This ensures that, for every question, the model has access to a consistent history of radiographic findings and EHR records as processed in the MIMIC-based clinical manifold.

The following Tabel \ref{tab:ehrxqa_stats} summarizes the filtering process and the resulting composition of the EHRXQA evaluation subset:

\begin{table}[h]
\centering
\resizebox{\columnwidth}{!}{
\begin{tabular}{lrr}
\hline \hline
Filtering Stage & Samples Removed & Remaining Samples \\ \hline
Initial Gold Test Set & - & 4,808 \\
Null Answer Removal & 258 & 4,550 \\
Non-Multimodal & 1,716 & 2,834 \\
Non-Independent & 334 & 2,500 \\
Modality Alignment Failures & 137 & 2,363 \\ \hline \hline
\end{tabular}
}
\caption{Statistics of the EHRXQA test set filtering process.}
\label{tab:ehrxqa_stats}
\end{table}

Regarding data splits, we use only the test set and further divide it according to answer type into binary and open-ended questions, resulting in 1,848 binary questions and 515 non-binary questions. This refined evaluation protocol requires the model to navigate the clinical hypergraph to associate specific EHR trajectories with radiographic evidence in order to generate accurate binary or open-ended answers.

\section{Discussion}

In this section, we analyze the functional mechanisms of HyperWalker's components and their contribution to diagnostic performance as substantiated by the experimental results.
\begin{figure*}
    \centering
    \includegraphics[width=\textwidth]{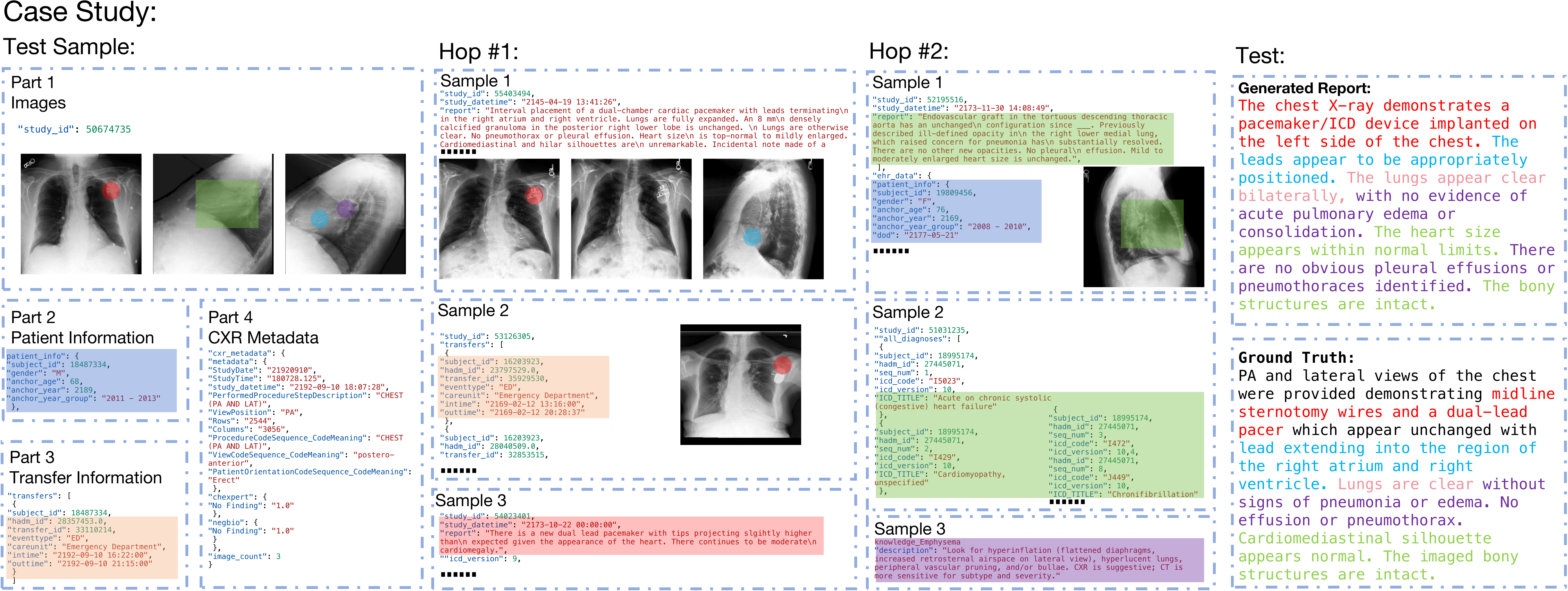}
    \caption{Case Study for MIMIC}
    \label{fig:case}
    \vspace{-\baselineskip}
\end{figure*}
\subsection{Relational Modeling and Manifold Construction via iBrochure}

The iBrochure module serves to transform independent data points into a structured clinical manifold, addressing the "sample-isolated inference" limitation. As shown in Table 4, the absence of EHR leads to a decline in BLEU-4 from 4.98 to 3.62. Functionally, this demonstrates that the hypergraph's ability to link longitudinal history with current imaging provides the necessary context that visual features alone lack. Furthermore, Table 1 reveals that while baseline models like Llama3.2-Vision and Qwen3-VL experience performance degradation or marginal gains when integrated with raw EHR data. However, HyperWalker maintains a superior F1 of 29.69. This validates that iBrochure’s redundancy-aware pruning and implicit hyperedge induction effectively distill salient clinical anchors from noisy longitudinal records.

\subsection{Directed Evidence Synthesis via Walker and RL Rewards}

The Walker agent replaces traditional attention-based retrieval with a reinforcement learning (RL) policy designed to navigate the hypergraph's high-dimensional space. The impact of this selection process is quantified in Table 4, where removing the accuracy reward ($\mathcal{R}_{\mathrm{acc}}$) reduces the F1 score to 23.56. This indicates that the agent’s scoring function is essential for identifying nodes that are semantically relevant to the specific patient state. Moreover, Table 2 highlights a significant efficiency gain: HyperWalker achieves higher generation quality (BLEU-4 4.98) in 4.66s, whereas Qwen3-VL-Thinking requires 86.56s to generate a reasoning chain. This confirms that "walking" to pre-existing evidence nodes is computationally more efficient than the autoregressive generation of "thinking" tokens, as the agent directly localizes diagnostic facts rather than hallucinating reasoning paths.

\subsection{Orthogonal Exploration via the Linger Mechanism}

The linger mechanism is designed to prevent "myopic" evidence accumulation by forcing the agent to explore complementary clinical facets. This is achieved through the orthogonalized query vector $z_{\mathrm{orth}}$, which steers the agent away from previously selected redundant regions. The necessity of this exploration is reflected in Table 4, where the removal of the diversity reward ($\mathcal{R}_{\mathrm{div}}$) drops the F1 score to 25.84. The mechanism's efficacy is most evident in the VQA results in Table 3, where HyperWalker achieves a state-of-the-art accuracy of 70.43\% on EHRXQA. Because VQA often requires bridging disparate data points (e.g., comparing a current X-ray with a historical lab result), the linger mechanism’s ability to move beyond initial similarity-based clusters allows for the synthesis of more comprehensive and diverse evidence.

\subsection{Adaptive Calibration via TTT}

HyperWalker utilizes TTT to bridge the gap between generalized pre-trained knowledge and individualized clinical evidence. By fine-tuning local adapters on retrieved clinical triplets during inference, the model calibrates its output distribution to the specific nuances of the test case. Table 2 shows that despite having a smaller parameter count ($\approx$ 4B), HyperWalker outperforms the 9B-parameter GLM-4.1V-Thinking in both BLEU-4 (4.98 vs 0.76) and F1 (29.69 vs 24.26). This suggests that local adaptation to the retrieved clinical subspace is more effective for precision diagnosis than relying on the static weights of a larger foundation model. The synergy between the multimodal fusion module and TTT ensures that the EHR-driven context is not merely appended, but actively modulates the model’s perception of visual features.

\section{Case Study}

In this section, we provide a qualitative analysis of HyperWalker's diagnostic trajectory to illustrate how the Walker agent navigates the clinical manifold to synthesize evidence. Fig. \ref{fig:case} presents a representative test case in which the model identifies complex findings by linking current visual evidence with longitudinal patient history. For clarity, regions of the same color across different images and textual components indicate corresponding clinical findings. Due to the extensive nature of clinical records, only the most relevant segments of retrieved evidence and patient data are shown. It should be noted that because the test samples are derived from fused features during inference, the annotations on these samples are inferred from the positions of the matched training samples in the hypergraph and do not represent the model’s raw output locations.

\subsection{Autonomous Evidence Navigation}

Upon receiving the initial test input (study 50674735), the Walker agent leverages the iBrochure manifold to retrieve historical context. The reasoning path is optimized through RL-based rewards that balance accuracy and diversity.

The agent’s retrieval trajectory demonstrates two key hops. In the first hop (Hop \#1), the agent identifies study 55403494 (Sample 1), which provides strong evidence of a dual-chamber cardiac pacemaker implantation with lead tips positioned in the right atrium and right ventricle. This finding aligns precisely with the visual evidence in the current X-ray image (highlighted in red), allowing the model to anchor the observed device clinically. Simultaneously, the retrieved visual features match study 53126305 (Sample 2) and correspond to the radiology report in study 54023401 (Sample 3). Moreover, the lateral chest image in the test sample shows lead positioning that matches study 55403494 (highlighted in light blue). The agent also detects that Sample 3 contains similar transfer information, revealing a latent relationship with the current case.

In the second hop (Hop \#2), the agent navigates to study 52195516 (Sample 1), which shows that the previously noted right lower medial lung opacity—once raising concerns for pneumonia—has largely resolved. By integrating longitudinal history from 51031235 (Sample 2), the agent confidently determines that the current pulmonary status is “clear” (highlighted in green), effectively avoiding misdiagnosis of residual shadows as active consolidation. Furthermore, the patient metadata in Sample 1 matches the test case. The agent also retrieves the medical knowledge node on “Emphysema” (Sample 3), which outlines key radiographic features such as hyperinflation and peripheral vascular rarefaction. By combining this knowledge with the observed pulmonary contour in the test sample (highlighted in purple), the agent successfully rules out acute pathology.

Overall, HyperWalker integrates these five critical samples along with domain-specific knowledge to construct a temporally and clinically coherent evidence chain. The final report demonstrates exceptional accuracy in pacemaker localization and lead positioning, while achieving high concordance with ground truth assessments of lung clarity and cardiac function.

Notably, across different hops, the agent discovers entirely new and complementary features, validating the effectiveness of the linger mechanism. This enables HyperWalker to mimic a human clinician’s iterative diagnostic reasoning, sequentially attending to salient findings while synthesizing longitudinal context. The resulting navigation exemplifies the agent’s ability to dynamically balance evidence retrieval and inference, forming a robust, cross-temporal diagnostic pathway.

\subsection{Report Synthesis and Clinical Grounding}

The final generated report exhibits high fidelity to both visual evidence and expert-written ground truth. The model correctly identifies the pacemaker on the left side of the chest (red) and confirms proper lead positioning. Furthermore, it accurately evaluates the heart size as normal (purple) and determines that the lungs are free of acute edema or consolidation (green).

Comparison with the ground truth confirms that HyperWalker effectively mitigates clinical hallucinations by grounding its reasoning in the structured clinical manifold. While baseline models often struggle with complex longitudinal histories, our Walker agent employs recursive TTT and orthogonal exploration to prioritize diagnostically salient anchors. This case study demonstrates that HyperWalker’s structured navigation leverages longitudinal EHRs alongside paired imaging evidence to construct evidence-based diagnostic pathways, replacing opaque chains of reasoning with transparent, clinically interpretable trajectories. Importantly, the agent’s capacity to discover complementary features and integrate medical knowledge allows it to simulate the nuanced decision-making process of experienced clinicians, supporting both accuracy and explainability in complex diagnostic tasks.

\section{Terminology Definitions}

This section formally defines the core terminology of the HyperWalker framework and elaborates on the functional roles of each component in simulating clinical diagnostic reasoning. The system is designed to construct a multimodal model that can rapidly adapt to clinical practice, leveraging a “deep diagnosis” mechanism to emulate physicians’ cognitive processes during diagnosis.

\subsection{HyperWalker}

HyperWalker is a multimodal medical vision–language model architecture proposed in this work. The framework integrates electronic health records (EHRs) with medical imaging data and adopts a lightweight fine-tuning strategy. Unlike conventional models that directly generate reports, HyperWalker performs sample-specific adaptive adjustment prior to inference. Its design objective is to simulate how physicians reason through complex cases by synthesizing multi-dimensional clinical evidence and prior experience, thereby enabling rapid adaptation to clinical practice.

\subsection{Deep Diagnosis}

“Deep Diagnosis” refers to a recursive fine-tuning process conducted before the inference stage. This process emulates the behavior of clinical experts who mentally review prior knowledge and clinical memory. By calibrating the model’s internal adapters prior to reasoning, HyperWalker transforms general medical knowledge into diagnostic strategies tailored to the current patient, effectively improving the clinical fidelity and logical coherence of the generated reports.

\subsection{iBrochure}

iBrochure is a hypergraph structure that implicitly stores clinical nodes and evidence, conceptualized as a medical handbook that supports large model reasoning. The letter i denotes its implicit nature. Serving as the knowledge backbone of HyperWalker, iBrochure organizes fragmented clinical events, such as laboratory measurements, imaging features, and diagnostic histories, into a knowledge manifold with high order associations. Rather than being a static database, it functions as a dynamic evidence space that captures complex relationships among medical entities and provides essential decision support for the model.

\subsection{HNSW: Hierarchical Retrieval Index}

To enable real-time localization within the large-scale iBrochure manifold, the system incorporates the Hierarchical Navigable Small World (HNSW) algorithm as its indexing mechanism. HNSW acts as a fast index for the “handbook,” allowing the model to rapidly identify the most relevant clinical clusters based on the semantic embedding of the current query. This mechanism ensures that the reasoning agent can efficiently lock onto critical pathological anchors, thereby improving the overall efficiency of diagnostic path construction.

\subsection{Walker}

Walker is the agent responsible for retrieval and path optimization within the iBrochure “handbook.” Its task is to search for the most relevant diagnostic evidence in the complex clinical manifold based on the patient’s initial findings. Each hop taken by Walker within the hypergraph represents an instance of clinical evidence retrieval and verification. This agent-based navigation paradigm ensures that the diagnostic process does not rely solely on opaque, weight-based model reasoning, but is instead grounded in traceable and comparable chains of clinical evidence, faithfully mirroring the cognitive pathway physicians follow from reviewing medical history to forming diagnostic conclusions.

\subsection{Linger Mechanism}

The Linger mechanism characterizes the gradual attenuation of effective information within fused multimodal features as traversal depth, measured in hops, increases across the clinical manifold. As the initial query signal naturally dissipates along the reasoning chain, Linger employs an orthogonal retrieval strategy to manage information decay. By explicitly targeting semantic residuals that are orthogonal to previously visited nodes, the mechanism ensures that the reasoning process explores complementary clinical directions rather than repeatedly sampling redundant information as the primary signal weakens.

\section{Limitations and Future works}
Despite the performance gains, several limitations warrant further investigation. Firstly, while HyperWalker incorporates longitudinal EHRs, it does not fully exploit the fine-grained temporal dependencies and sequential dynamics inherent in disease progression, potentially overlooking the significance of the intervals between clinical events. Secondly, the construction and maintenance of the multimodal hypergraph, although effective, remain computationally intensive in terms of both time and memory overhead when scaling to massive, real-time clinical databases. Thirdly, the multi-hop reasoning process currently relies on the Walker agent's navigation and orthogonal decomposition vectors, yet it lacks the internal "chain-of-thought" or intermediate reflective reasoning typically found in large-scale vision-language models during the multihop steps. Building upon these observations, future research will aim to refine the algorithmic framework to address these constraints while extending its applicability to a broader spectrum of clinical modalities to support more comprehensive cross-departmental diagnostics.

\end{document}